\documentclass[10pt,twocolumn,letterpaper]{article}

\usepackage[pagenumbers]{cvpr} 

\usepackage{graphicx}   
\usepackage{booktabs}   
\usepackage{multirow}   
\usepackage{caption}    

\usepackage{pifont}    

\newcommand{\cmark}{\textcolor{green}{\ding{51}}} 
\newcommand{\xmark}{\textcolor{red}{\ding{55}}} 

\definecolor{cvprblue}{rgb}{0.21,0.49,0.74}
\usepackage[pagebackref,breaklinks,colorlinks,allcolors=cvprblue]{hyperref}

\usepackage{colortbl}
\usepackage{xcolor} 
\usepackage{tcolorbox}

\def\eg{\emph{e.g.}\xspace}

\newtcolorbox[list inside=prompt,auto counter,number within=section]{prompt}[1][]{
    colbacktitle=black!60,
    coltitle=white,
    fontupper=\footnotesize,
    boxsep=5pt,
    left=0pt,
    right=0pt,
    top=0pt,
    bottom=0pt,
    boxrule=1pt,
    title={#1},
    #1, 
}

\title{\includegraphics[width=1.5em]{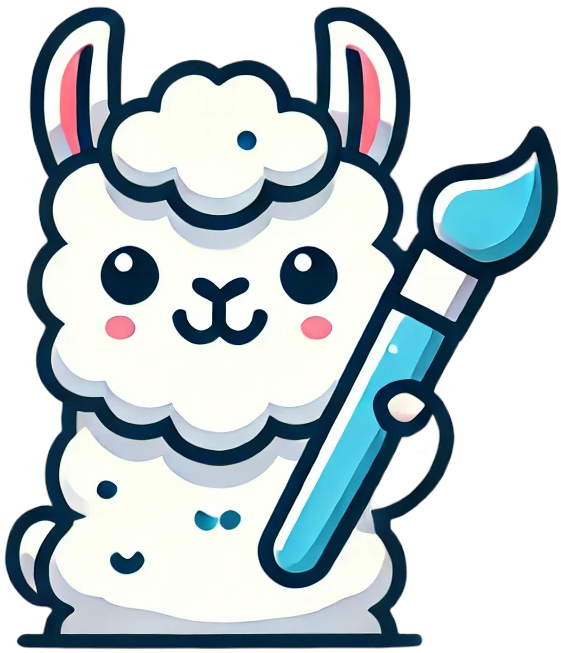}ChatGen: Automatic Text-to-Image Generation
From FreeStyle Chatting}

\author{Chengyou Jia$^{1}$\footnotemark[1],\; Changliang Xia$^{1}$\footnotemark[1],\; Zhuohang Dang$^{1}$,\; Weijia Wu$^{2}$,\; Hangwei Qian$^{3}$,\; Minnan Luo$^{1}$\footnotemark[2] \\
$^{1}$Xi’an Jiaotong University \; $^{2}$National University of Singapore\quad $^{3}$CFAR, A*STAR \\
\tt\small {cp3jia}@stu.xjtu.edu.cn  
}
\begin{document}

\maketitle

\renewcommand{\thefootnote}{\fnsymbol{footnote}} 
\footnotetext[1]{Equal Contribution.}
\renewcommand{\thefootnote}{\fnsymbol{footnote}} 
\footnotetext[2]{Corresponding author.}

\begin{abstract}

Despite the significant advancements in text-to-image (T2I) generative models, users often face a trial-and-error challenge in practical scenarios.
This challenge arises from the complexity and uncertainty of tedious steps such as crafting suitable prompts, selecting appropriate models, and configuring specific arguments, making users resort to labor-intensive attempts for desired images.
This paper proposes \textbf{Automatic T2I} generation, which aims to automate these tedious steps, allowing users to simply describe their needs in a freestyle chatting way.
To systematically study this problem, we first introduce \textbf{ChatGenBench}, a novel benchmark designed for Automatic T2I. 
It features high-quality paired data with diverse freestyle inputs, enabling comprehensive evaluation of automatic T2I models across all steps.
Additionally, recognizing Automatic T2I as a complex multi-step reasoning task, we propose \textbf{ChatGen-Evo}, a multi-stage evolution strategy that progressively equips models with essential automation skills. 
Through extensive evaluation across step-wise accuracy and image quality, ChatGen-Evo significantly enhances performance over various baselines. Our evaluation also uncovers valuable insights for advancing automatic T2I. All our data, code, and models will be available in \url{https://chengyou-jia.github.io/ChatGen-Home}

\end{abstract}    
\vspace{-0.3cm}
\section{Introduction}
\label{sec:intro}

In recent years, text-to-image (T2I) generative models have attracted considerable attention~\cite{nichol2022glide,ramesh2022hierarchical,betker2023dalle3,rombach2022high}.
Building on advancements in large-scale T2I models such as DALL-E \cite{ramesh2022hierarchical} and Stable Diffusion~\cite{rombach2022high}, the open-source community has significantly expanded the capabilities of T2I generation.
Researchers fine-tune open-source models on specialized datasets, resulting in a diverse selection of task-specific models available on platforms like Civitai~\cite{2022civitai} and Hugging Face~\cite{2016hugggingface}.
This variety provides users with a broader range of options to meet customization needs, facilitating the growing adoption of T2I models in real-world applications.

\begin{figure}[!t]
  \centering
   \includegraphics[width=0.92\linewidth]{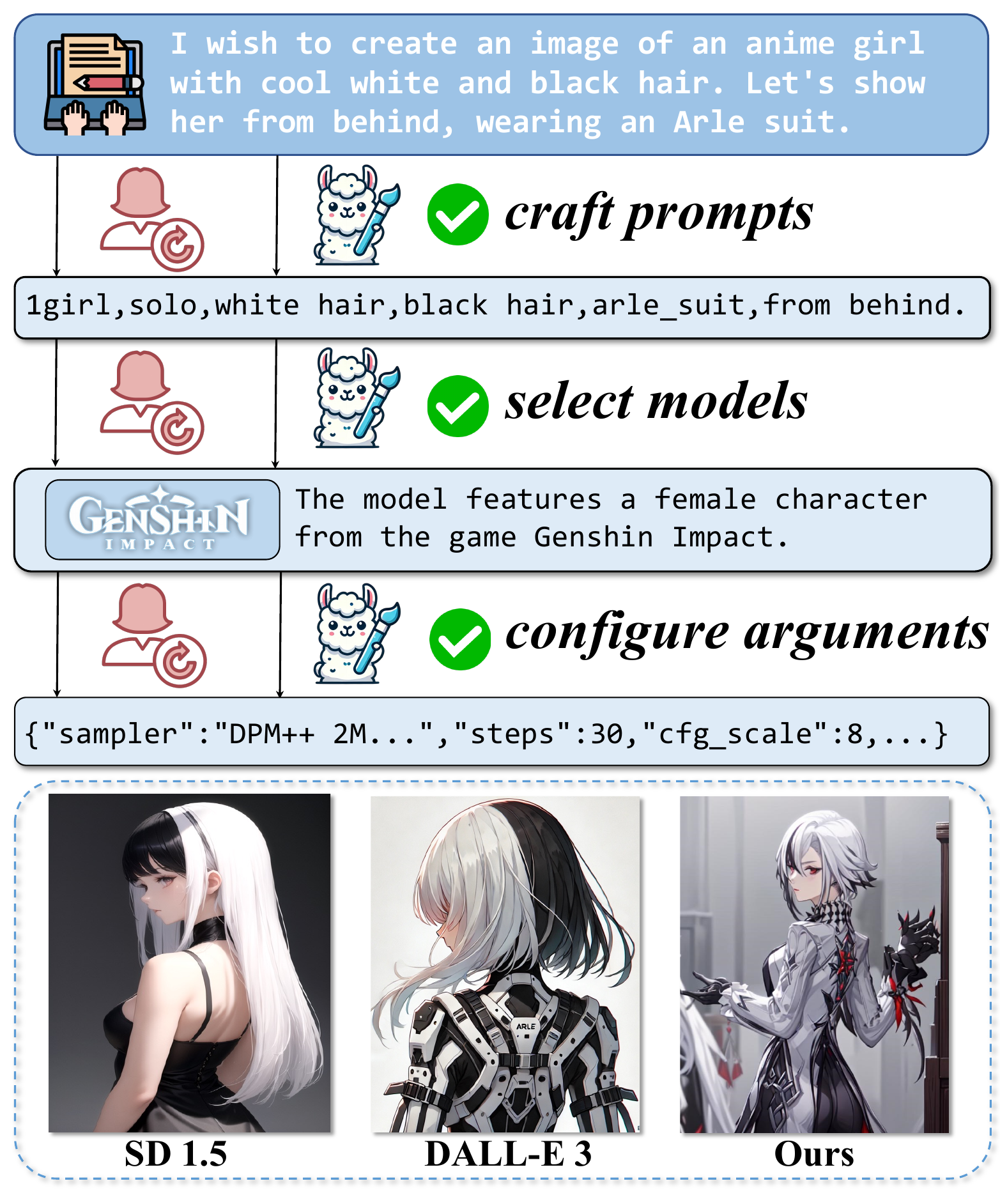}
   \vspace{-4mm}
   \caption{Illustration of tedious steps in T2I. Our method can select an appropriate model with suitable prompts and arguments. \textit{Note: Arle is a character from the game Genshin Impact.}}
   \label{fig:intro}
   \vspace{-8mm}
\end{figure}

However, the rapid development of T2I models within the open-source community has also introduced significant challenges for users.
When non-experts attempt to create images with specific requirements, they often encounter a trial-and-error process involving several tedious steps. As shown in Figure \ref{fig:intro}, these steps include \textbf{crafting suitable prompts}, \textbf{selecting appropriate models}, and \textbf{configuring specific model arguments}. The complexity and uncertainty of each step turn the process into an arduous journey, resembling ``mice in a maze". In real-world scenarios, this iterative process consumes substantial time and resources as users continuously adjust settings to regenerate images. 
Therefore, we pose the challenging problem: \textbf{\textit{can we automate these labor-intensive steps in T2I generation, allowing users to simply describe their needs in a chatting style and receive desired images effortlessly?}}

Previous attempts, such as BeautifulPrompt~\cite{cao2023beautifulprompt,wang2024diffchat} for generating high-quality prompts from low-quality ones and DiffAgent~\cite{zhao2024diffagent} for model selection using large language models, have made some progress in addressing these challenges.
However, these methods only seek to automate a specific step in Figure \ref{fig:intro}, lacking comprehensive research on the automation of entire T2I process. Moreover, they fail to support diverse types of freestyle input, leaving them far from real-world scenarios.
To bridge these gaps, we aim to develop the model that can accept arbitrary user input, similar to ChatGPT~\cite{openai2024chatgpt}, and automatically generate all necessary components for generation, termed as \textbf{Automatic T2I}.

To this end, we first introduce \textbf{ChatGenBench}, the benchmark specifically designed for this task.
ChatGenBench offers a substantial dataset of high-quality paired data from $6,807$ customized models.
Each data pair comprises a user’s freestyle chat input, a refined prompt, an appropriate model, and optimal arguments.
This comprehensive step-by-step trail enables step-wise evaluation of automatic T2I models, ensuring both quality assessment of the final image output and precise identification of potential automation bottlenecks.
Furthermore, ChatGenBench integrates various types of data, enabling testing with multimodal and historical input to simulate practical scenarios.

An intuitive approach to achieve the goal is to build supervised fine-tuning (SFT) data and tune multimodal large language models.
However, we argue that automated T2I should be viewed as a complex multi-step reasoning problem.
Directly predicting outputs leads the model to focus on simple text mappings, rather than developing the diverse skills needed for robust automation.
Inspired by OpenAI’s o1 \cite{openai2024learning}, we propose \textbf{ChatGen-Evo} with a multi-stage evolution strategy. 
This approach provides targeted feedback at each stage, progressively equipping models with essential automation skills. 
First, we construct SFT data to train the model to generate high-quality prompts from freestyle inputs. 
Next, we augment the model’s vocabulary with specialized ModelTokens, enabling effective model selection without affecting other functions.
Finally, we guide the model in configuring arguments based on prompts and models selected in previous stages. By decomposing the task into clear stages, the model gains crucial automation skills, delivering outputs aligned with user expectations.

We conduct a comprehensive evaluation of various methods on the novel ChatGenBench to study Automatic T2I. ChatGen-Evo significantly outperforms other methods across all metrics, including both step-wise assessments and image quality evaluations. Extensive quantitative and qualitative results, along with ablation studies, underscore the critical importance of the multi-stage evolution strategy. Additionally, further experiments reveal the impact of different stages on the final T2I output, uncovering valuable insights into the challenges and opportunities in Automatic T2I. We highlight some promising directions, such as exploring the scaling laws of prompt rewriting, improving model selection in few-shot scenarios, and leveraging reasoning approaches to enhance performance. The contributions of this paper are summarized as follows:

\begin{itemize}[leftmargin=*]
\item We propose the novel and challenging problem of Automatic T2I generation, aiming to develop the model that can handle user's freestyle chatting and automatically produce all necessary components for image generation.

\item We introduce ChatGenBench, a benchmark specifically designed for Automatic T2I.
It includes a comprehensive step-by-step trail for step-wise evaluation, supporting multimodal or historical user inputs.

\item We present ChatGen-Evo, which adopts a multi-stage evolution strategy. By decomposing the task into distinct stages, ChatGen-Evo enables the model to progressively acquire essential Automatic T2I capabilities.

\item Extensive experiments on ChatGenBench not only demonstrate the superiority of ChatGen-Evo but also provide valuable insights for advancing automatic T2I.

\end{itemize}

\section{Related Works}
\label{sec:related}

\begin{figure*}[!t]
  \centering
   \includegraphics[width=0.9\linewidth]{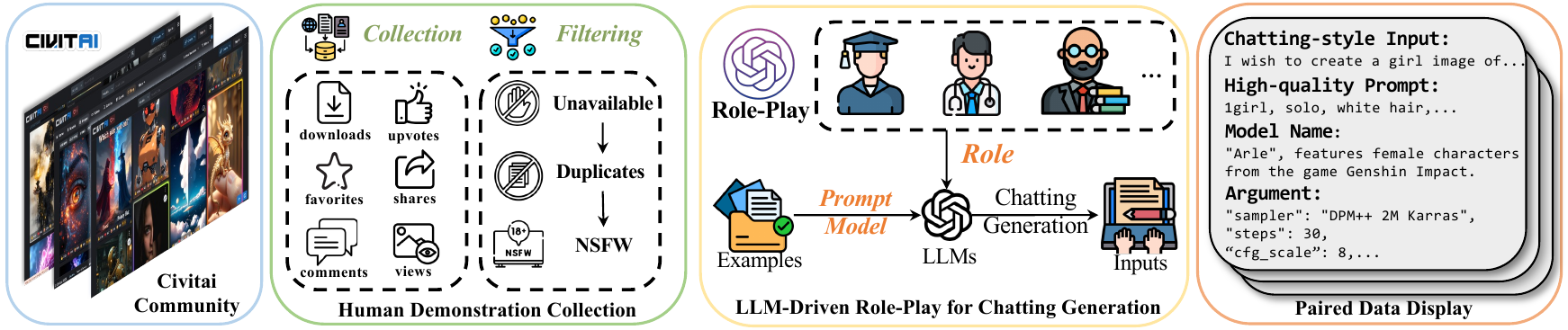}
   \vspace{-0.2cm}
   \caption{Illustration of the data collection pipeline.}
   \label{fig:datset}
   \vspace{-6mm}
\end{figure*}

\subsection{Text-to-Image Generation}

With the advancement of diffusion models~\cite{ho2020denoising,song2020denoising,peebles2023scalable}, text-to-image (T2I) generation~\cite{nichol2022glide,ramesh2022hierarchical,betker2023dalle3} has demonstrated exceptional capabilities in high-quality image generation and textual description alignment. Large-scale models such as DALL-E~\cite{betker2023dalle3} enhance text-image alignment by leveraging the joint feature space of CLIP~\cite{radford2021learning}. Moreover, Stable Diffusion~\cite{rombach2022high}, a well-established open-source model, has gained substantial attention. Numerous methods have been developed to fine-tune it or design additional modules for specialized tasks, such as customized image generation~\cite{ruiz2023dreambooth,kumari2023multi,li2024blip}, layout-to-image generation~\cite{jia2024ssmg,lian2023llm,li2023gligen,zhang2023adding} and image edit~\cite{brooks2023instructpix2pix,kawar2023imagic}. These diverse models have significantly expanded the capabilities of T2I. However, they also present significant learning challenges for non-expert users, underscoring the increasing need for automatic T2I generation.

\subsection{LLMs for Text-to-Image Generation}

Large language models (LLMs), like ChatGPT~\cite{openai2024chatgpt}, Llama~\cite{touvron2023llama}, have demonstrated impressive capability in language understanding~\cite{brown2020language} and problem solving~\cite{yao2022react}. Recently, LLMs have also begun to be applied to image generation. Recently, LLMs have also been applied to image generation.~\cite{lian2023llm,dong2023dreamllm,yang2024mastering,feng2024ranni,omost} leverage LLMs to generate detailed layout information from complex prompts, enabling the creation of sub-elements and control over their positioning. ~\cite{qin2024diffusiongpt,wang2024genartist,zhao2024diffagent} propose employ LLMs for model selection, akin to tool usage~\cite{patil2023gorilla,yang2024gpt4tools,qin2023toolllm}. 
However, the above methods still require specialized prompt inputs and involve complex model usage. Currently, no approach thoroughly considers leveraging LLMs to relieve users from all tedious steps in T2I. Our method aims to address this gap.

\section{Methodology}

Our goal is to relieve users from tedious steps and automate to produce the desired images from user's freestyle input. 
To achieve this objective, we first introduce ChatGenBench, a benchmark comprising a large set of user inputs in a chatting style, built upon a foundation of over 6000 personalized models to evaluate automated image generation results.
A comprehensive description of data collection, construction, and analysis is provided in \cref{sec:chatgenbench}. 
Then, we present ChatGen-Evo in \cref{sec:chatgenevo}, which uses a multi-stage evolution strategy to train MLLM for Automatic T2I.

\subsection{ChatGenBench: Benchmarking Automatic T2I}
\label{sec:chatgenbench}

For clarity, an example of this data is shown on the right side of Figure \ref{fig:datset}. The following sections detail the data collection and construction process, which primarily involves \textbf{High-Quality Human Demonstration Collection} and using \textbf{LLM-driven Role-Play} to simulate user input.

\begin{table*}[t]
\centering
\caption{
Comparison of different methods for benchmarking automatic text-to-image generation. 
}
\vspace{-3.5mm}
\label{tab:bench_compare}
\begin{tabular}{@{}cccccccc}
\toprule
& \shortstack{TrainSet \\(\# Numbers)} & \shortstack{TestSet \\(\# Numbers)} & \shortstack{Based \\Models?} & \shortstack{Step-wise \\ Evaluation?} & \shortstack{Freestyle \\User-input?}  & \shortstack{Multi-modal \\ Support?} & \shortstack{History\\Support?}  \\ 
\midrule
BeautifulPrompt~\cite{cao2023beautifulprompt} &143K & 2K & Single & \xmark & \cmark & \xmark & \xmark \\
DiffChat~\cite{wang2024diffchat} &234K  & 5K & Single & \xmark & \cmark & \xmark & \cmark \\

DiffusionGPT~\cite{qin2024diffusiongpt} &-  & - & Multi($\approx$20) & \xmark & \cmark & \xmark & \xmark \\

DABench~\cite{zhao2024diffagent} & 40K & 5K & Multi(5K) & \xmark & \xmark & \xmark & \xmark \\

\midrule
\textbf{ChatGenBench} & 256K & 14K & Multi(6K) & \cmark & \cmark & \cmark & \cmark \\
\bottomrule
\end{tabular}
\vspace{-6mm}
\end{table*}

\subsubsection{High-Quality Human Demonstration Collection}

Civitai~\cite{2022civitai} is a vibrant community where users share customized AI models for generating high-quality images. Members contribute demonstrations with detailed prompts, model specifications, and arguments, supported by a feedback system that ensures quality validation. These features make Civitai an ideal platform for collecting raw data.

\noindent \textbf{Collection:} We start by collecting demonstrations based on established evaluation metrics within Civitai, including download counts, upvotes, and other user feedback. These metrics enable us to collect data that has been validated through community engagement. By focusing on these indicators, we identify a subset of high-quality results.

\noindent \textbf{Filtering:} Following the initial collection, we implement a rigorous filtering process to ensure data quality. This involves excluding demonstrations associated with inactive or outdated models, removing duplicates, and filtering out NSFW content.  This careful curation refined the dataset to include the most effective demonstrations. Ultimately, this process results in a curated set of 44,881 high-quality human demonstrations across 6,807 unique models.

\subsubsection{LLM-Driven Role-Play for Chatting Generation}

While the demonstrations collected from communities include the essential procedural information needed for automation, they lack a critical element: freestyle chatting inputs. Such data is not available on open platforms, which has been a key limitation for previous methods~\cite{zhao2024diffagent}. 
To address this, we propose an LLM-Driven Role-Play strategy for Chatting Generation. As shown in Figure \ref{fig:datset}, we predefine over 100 roles from everyday life (\eg, students, doctors, professors) and prompt the LLM to simulate these roles, translating each demonstration into freestyle chatting input with tones and habits of the character. This approach significantly enhances data diversity and creates more lifelike inputs. To further enhance reverse synthesis diversity, we employed multiple versions of (M)LLMs to complete all generation tasks, displayed in Table \ref{tab:mllm}. Additionally, we set the temperature parameter to 0.9 to make the model more stochastic. We also utilized BertScore~\cite{zhangbertscore} to filter out results with a similarity greater than 0.8.

Moreover, we define three types of freestyle input formats: single-input, consisting of a single chatting-style sentence; multimodal-input, combining a sentence with an image; and history-input, comprising multiple rounds of dialogue history. These formats effectively simulate how users typically inquire about image generation needs, greatly expanding the practical value of automatic T2I.  An example of the single-input prompt is shown in the following.

\begin{table}[ht]
\centering
\caption{Overview of MLLMs utilized in chatting generation. The table lists different MLLMs on the left, with corresponding approximate numbers of generated chatting queries on the right.}
\vspace{-0.2cm}
\begin{tabular}{cc}
\toprule
\textbf{MLLM} & \textbf{Numbers} \\
\midrule
gpt-4o-2024-08-06 & $\approx$100,000\\
gpt-4o-mini-2024-07-18 & $\approx$70,000\\
gpt-4-turbo-2024-04-09 & $\approx$70,000\\
claude-3-5-sonnet-20240620 & $\approx$100,000\\
\bottomrule
\end{tabular}
\vspace{-0.2cm}
\label{tab:mllm}
\end{table}

\begin{prompt}[title={Prompt: Sing-Input chatting generation}]
\label{prompt:1}
\textbf{System:} You are a professional user experience designer who plays various personas to convert complex and professional content for non-professional users. Please merge the following prompt and model information into a single freestyle query. Remove any obvious details that non-professional users would avoid. Make it similar to what non-professional users may write. The converted single-text query should be colloquial and as brief as possible.

\par 
\textbf{Require:} \{ROLE\},  \{PROMPT\}, \{MODEL\}, \{Example 1,...,n\}

\par 
\textbf{Input:} You are playing the \{ROLE\}. Please generate a single text query based on the following professional \{PROMPT\} for the \{MODEL\}. You can refer to following examples: \{Example 1,...,n\}.
\end{prompt}

\subsubsection{Benchmark Construction}

Following the above steps, we generate 330,970 freestyle inputs from 44,881 demonstrations. Considering the large-scale generated data, we don't randomly split a test set for benchmarking. Instead, we carefully select high-quality, non-overlapping samples as the benchmark for evaluation. The selection involves the following steps:

\noindent \textbf{TestSet Split:} We split the data based on model origin. For data associated with the same model, we use BERTScore to assess the similarity between samples. From this, we select about 20\% most semantically distinct data as initial TestSet in Table \ref{tab:stat}. This maximizes non-overlap with TrainSet.

\noindent \textbf{TestSet Filtering:} Building on the initial TestSet, we perform multiple filtering rounds to ensure the final ChatGenBench's quality. This involves the following steps:
\begin{itemize}[leftmargin=*]
\item \underline{\textit{Length Filtering}}: Remove excessively long entries or those with too many dialogue turns for consistency.
\item \underline{\textit{Colloquialism Check}}: Utilize the Spacy~\cite{spacy} module to filter data for alignment with natural, everyday language.
\item \underline{\textit{LLM-Based Evaluation}}: Employ LLM to assess and select data that best matches the chatting tone.
\item \underline{\textit{Manual Verification}}: Manual verification is conducted to eliminate any inappropriate or irrelevant samples.
\end{itemize}

Through this rigorous process, we refine the initial 74,364 entries to a final set of 14,564 high-quality, freestyle chatting samples, constituting the novel benchmark.

\noindent \textbf{Setting Division:} In practical scenarios, new models often have limited demonstration data, underscoring the importance of evaluations under constrained data. Therefore, we further divide ChatGenBench into two settings—Supervised and Few-shot—based on the availability of samples for each model in the TrainSet. The complete data composition is detailed in Table \ref{tab:stat}.

\begin{table}[h!]
  \centering
  \caption{Summary of dataset statistics.}
  \vspace{-2mm}
  \begin{tabular}{lccccc}
    \toprule
    Dataset & Total & Single & M-Modal & History \\
    \midrule
    TrainSet  & 256,606 & 147,888 & 69,548 & 39,170 \\
    TestSet Init  & 74,364 & 42,838 & 20,214 & 11,312 \\
    \midrule
    Benchmark & 14,564 & 11,011 & 1,668 & 1,132 \\
    \midrule
    Supervised & 10,240 & 8,009 & 1,099 & 1,132 \\
    Few-Shot  & 4,324 & 3,002 & 569 & 753 \\
    \bottomrule
  \end{tabular}
  \vspace{-6mm}
  \label{tab:stat}
\end{table}

\subsubsection{Benchmark Analysis}
ChatGenBench offers distinct advantages over previous benchmarks, as summarized in Table \ref{tab:bench_compare}. On the data level, ChatGenBench includes large-scale high-quality data and a broader range of T2I models. Additionally, human demonstrations provide relative ground truth across each step, allowing for step-wise evaluation that pinpoints potential challenges in automation models. ChatGenBench is also the novel benchmark to support multiple input types, making it more aligned with real-world scenarios.

\begin{figure*}[!t]
  \centering
   \includegraphics[width=0.9\linewidth]{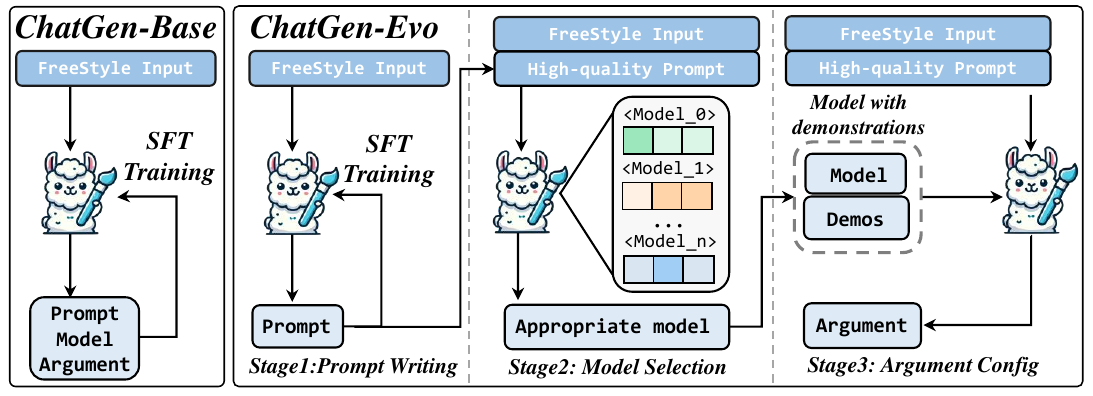}
   \vspace{-2mm}
   \caption{Illustration of the framework for ChatGen-Base and ChatGen-Evo.}
   \label{fig:method}
   \vspace{-5mm}
\end{figure*}

\subsection{ChatGen: Achieving Automatic T2I}
\label{sec:chatgenevo}

Our goal is to train the model capable of processing freestyle user inputs and generating the necessary components for image generation (prompt, model, and argument), thereby achieving Automatic T2I. In this section, we first introduce a baseline method, ChatGen-Base. We then analyze the limitations of this approach and subsequently propose ChatGen-Evo, which leverages the multi-stage evolution strategy to enhance performance.

\subsubsection{ChatGen-Base with SFT}

We first apply supervised fine-tuning (SFT) to develop the ChatGen-Base model as an intuitive baseline. Given a set of freestyle chatting inputs $c$ (which may include text, images, and historical context) and their corresponding outputs comprising prompt $p$, model $m$, and argument $r$, we use the auto-regressive objective to maximize the following loss:

\begin{equation}
\label{equ:sft_baseline}
\centering
L_{sft}^{base} = -\sum_{t} \log P_{\pi}(p, m, r \mid c, *_{<t}). 
\end{equation}

ChatGen-Base satisfies the essential requirements for generating outputs, including handling diverse input modalities and producing corresponding responses. However, even after fine-tuning SOTA open-source MLLMs, the model frequently produces unsatisfactory results. 

We attribute this to the complexity of automatic T2I, which requires multi-step reasoning. The quality of generated prompts directly impacts model selection, while model selection serves as a prerequisite for argument configuration. 
Previous studies~\cite{wei2022chain,openai2024learning,lightman2023let} have shown that directly predicting answers often fails in multi-step reasoning tasks. Additionally, supervised fine-tuning tends to encourage the model to learn simple text mappings instead of developing skills needed for automation, leading to poor generalization.

\subsubsection{ChatGen-Evo}

To address the limitations of ChatGen-Base, we propose ChatGen-Evo, which trains MLLM $M$ using a multi-stage evolution strategy. Instead of relying on final outcome supervision as in traditional fine-tuning, the multi-stage evolution strategy in ChatGen-Evo employs stage supervision. By providing more precise feedback at each stage, this approach gradually enables the MLLM to acquire the necessary capabilities for automated T2I. As shown in Figure \ref{fig:method}, the training process includes three main stages:
\vspace{-0.4cm}
\paragraph{Stage 1: Prompt Writing via SFT.}
We first train the MLLM using SFT with pairs of freestyle inputs and high-quality prompts. Different from the objective in ChatGen-base, the current stage focuses on a more specific and simplified task instead:
\begin{equation}
L_{sft}^{stage1} = -\sum_{t} \log P_{\pi}(p \mid c', *_{<t}).
\end{equation}
Here, \( c' \) represents the freestyle input with a prompt prefix that clarifies the task being performed. This prefix helps preserve the MLLM’s original capabilities, minimizing catastrophic forgetting~\cite{luo2023empirical}. Through this stage, the model learns how to rewrite inputs into effective prompts. 

\vspace{-0.4cm}
\paragraph{Stage 2: Model Selection via ModelToken.}

We introduce the ModelToken strategy to equip the model with model selection capabilities without impacting the prompt-writing ability learned in Stage 1. Inspired by token learning~\cite{hao2024toolkengpt,jia2024agentstore}, ModelToken extends this approach by representing each candidate T2I model as a unique token within the MLLM’s vocabulary. Specifically, the model tokens are parameterized as an embedding matrix \( W_{\mathcal{M}} \in \mathbb{R}^{|\mathcal{M}| \times d} \) and appended to the original word token matrix \( W_{\nu} \in \mathbb{R}^{|\mathcal{V}| \times d} \). 

\textbf{\textit{ModelToken Training:}} during training, the user input \( c \) and prompt \( p \) are concatenated as a prefix, with the special model token \(\texttt{<Model}\_i\texttt{>}\) appended as the ground truth for next-token prediction. The training objective is:
\begin{equation}
L(W_{\mathcal{M}}) = -\log P_{\pi}(\texttt{<Model\_i>} | c,p).
\end{equation}
Unlike typical next-token prediction training, the embedding matrix \( W_{\mathcal{M}} \) represents the only tunable parameters, significantly enhancing training efficiency. With a small parameter size, fewer training samples are required, improving performance in scenarios with limited data.

\textbf{\textit{Inference for Model Selection:}} Once the embedding matrix is trained, the inference process concatenates the model token and original word token, forming the new language modeling head of the MLLM. In this way, the MLLM predicts the next token with the following probability:
\begin{equation}
\vspace{-0.1cm}
P_{\pi}(m |c,p) = \text{softmax}([W_{\nu}; W_{\mathcal{M}}] \cdot h_{i-1}),
\vspace{-0.1cm}
\end{equation}
where the operation \(\left[;\right]\) denotes concatenation, and \( h_{i-1} \in \mathbb{R}^d \) represents the last hidden state. Once a model token is predicted, the MLLM stops decoding, and the corresponding model \( m \) is selected. Additionally, information such as the model's description and demonstrations is loaded for subsequent use. After Stage 2, the model not only retains its prompt-writing skills but also learns to select models.

\vspace{-0.4cm}
\paragraph{Stage 3: Argument Configuration via In-Context Learning.} After the above two stages, we have obtained the prompt \( p \) and model \( m \) from the original user input \( c \). The model now needs to generate the appropriate argument configuration to complete the final generation. Due to the careful design of the previous stages, the model’s in-context learning ability is maximally preserved. Therefore, we adopt a training-free approach: we guide the MLLM using in-context demonstrations of \( m \), the user’s original input \( c \), and the rewritten prompt \( p \). The MLLM can follow the demonstration pattern to complete the parameter configuration for the current user request:
\vspace{-0.2cm}
\begin{equation}
\centering
a = M(c, p, D(m)),
\vspace{-0.2cm}
\end{equation}
where $D(m)$ represents the set of demonstrations for model $m$ in the training dataset. Thanks to the prior acquisition of the relevant model and prompt in earlier stages, this approach frees up context space, enabling extensive demonstrations. Additionally, the train-free approach avoids interfering with the trained model from the previous two stages.


\section{Experiment}
\label{sec:experiment}

We conduct a comprehensive evaluation of various methods on the novel ChatGenBench. First, in section \ref{exp:results}, we compare ChatGen-Evo with other baseline models, highlighting the efficacy and efficiency of our multi-stage evolution strategy. Next, in section \ref{exp:analysis}, we perform extensive ablation studies, uncovering the impact of each step on the final results and providing valuable insights. Finally, we provide visualizations of the generated images in section 
 \ref{exp:visualization}.

\subsection{Experimental Settings}

\paragraph{Training Setups.}
\label{exp:setting}

We adopt InternVL2~\citep{chen2024far} as the base MLLM, fully fine-tuning it for both ChatGen-Base and the first stage of ChatGen-Evo. In the second stage of ChatGen-Evo, all model parameters are frozen except for the ModelToken embeddings. We employ the AdamW optimizer with a learning rate of 4e-5 and a weight decay of 1.0 over 5 epochs, maintaining these settings consistently across training stages. All experiments are conducted on 8 A100 GPUs. 

\vspace{-0.3cm}
\paragraph{Metrics for Step-wise Evaluation:} Leveraging the comprehensive process data in ChatGenBench, we introduce the step-wise evaluation metrics to assess the distinct abilities of automatic T2I models in key stages:

\begin{itemize}[leftmargin=*]
\item \textbf{\textit{Prompt BERTScore:}} To assess prompt rewriting ability, we use BERTScore~\cite{zhangbertscore} to compare predicted prompts with high-quality, human-validated prompts. BERTScore leverages pre-trained contextual embeddings to match words in candidate and reference sentences. The metric ranges from 0 to 1, with 1 indicating highly similar meanings and 0 signifying complete dissimilarity.

\item \textbf{\textit{Selection Accuracy:}} We calculate the accuracy of model selection by comparing the predicted T2I model with the human-validated model.

\item \textbf{\textit{Argument Accuracy:}} We evaluate argument configuration by calculating the exact match accuracy between the predicted arguments and the human-validated arguments. The overall argument accuracy is obtained by averaging the accuracy across all individual arguments.

\end{itemize}
It is important to note that obtaining the absolute ground truth across all stages is nearly impossible due to the infinite search space. Therefore, we use human-validated high-quality records as \textbf{relative ground truth}. While these are not absolute, the comprehensive evaluation of the large-scale benchmark is able to reflect the capabilities of automation models, as confirmed by experimental results.
\vspace{-0.3cm}
\paragraph{Image Quality Evaluation:}  We use HPS v2~\cite{wu2023human} and ImageReward~\cite{xu2023imagereward} metrics to assess the quality of generated images, reflecting alignment with human preferences. Additionally, we employ FID and CLIP Score to evaluate how well the generated images meet user requirements. FID~\cite{fid} measures the distance between automatically generated images and human-validated high-quality images, while CLIP Score~\cite{hessel2021clipscore} calculates the similarity between the generated images and human-validated prompts. To provide an intuitive and comprehensive measure of image quality, we normalize and combine these four metrics into an aggregated score, \textbf{Unified Metric}~\cite{zhao2024diffagent}. Each of above
scores are normalized to the range [0,1] and the final score is computed by averaging the value:
\begin{equation}
\begin{aligned}
    S_{unified} = &\frac{1}{4}
      \left.( (1 - norm(S_{fid})) +  norm(S_{clipscore})  \right.\\
      &\left. + norm(S_{hps}) + norm(S_{reward}) \right.).
\end{aligned}
\end{equation}
By integrating these diverse metrics, the Unified Metric offers a more nuanced and comprehensive understanding of image quality, facilitating more informed comparisons and evaluations of generated images.

\begin{table*}[!ht]
  \centering
  \caption{The Step-wise and Final evaluation results of different methods on ChatGenBench.}
  \vspace{-3.5mm}
  \captionsetup{skip=5pt}
  \small
    \begin{tabular}{llccc|ccccc}
    \toprule
        && \multicolumn{3}{c|}{\textbf{Step-wise Evaluation}} & \multicolumn{5}{c}{\textbf{Final Evaluation}} \\
     && \begin{tabular}[c]{@{}c@{}}Prompt\\BertScore $\uparrow$\end{tabular}      & \begin{tabular}[c]{@{}c@{}}Selection\\Acc $\uparrow$\end{tabular}  & \begin{tabular}[c]{@{}c@{}}Config\\Acc $\uparrow$\end{tabular} & \begin{tabular}[c]{@{}c@{}}FID\\Score $\downarrow$\end{tabular}      & \begin{tabular}[c]{@{}c@{}}CLIP\\Score $\uparrow$\end{tabular}  & \begin{tabular}[c]{@{}c@{}} HPS\\v2 $\uparrow$ \end{tabular}        & \begin{tabular}[c]{@{}c@{}} Image\\Reward $\uparrow$ \end{tabular}& \begin{tabular}[c]{@{}c@{}} Unified\\Metric $\uparrow$ \end{tabular} \\
    \midrule
        \multirow{5}{*}{Supervised}
        & Baseline               &0.026           &-          &-           &32.7           &64.6           &20.2           & -34.6 & 37.3    \\
        &ChatGen-Base(2B)           &0.184           &0.206          &0.384           &21.3          &69.9          &23.5          & 2.4   & 59.0  \\
        &ChatGen-Base(4B)           &0.197           &0.230          &0.490           &20.7         &70.0          &23.4          & 1.5    & 58.7 \\
        &ChatGen-Base(8B)           &0.208           &0.264          &0.509           &20.8          &70.7          &23.9          & 4.0    & 60.7 \\
        &\cellcolor{gray!22}ChatGen-Evo (2B)           &\cellcolor{gray!22}0.247           &\cellcolor{gray!22}0.328          &\cellcolor{gray!22}0.537           &\cellcolor{gray!22}19.1              &\cellcolor{gray!22}72.9              &\cellcolor{gray!22}25.1              &\cellcolor{gray!22}8.9     & \cellcolor{gray!22}65.9     \\
    \midrule
        \multirow{5}{*}{Few-shot}
        & Baseline               &0.055           &-          &-           &54.4          &63.4          &20.0          & -40.2  & 29.7   \\
        &ChatGen-Base(2B)           &0.221           &0.153          &0.349           &42.8          &69.1          &23.3          & -4.8   & 51.1  \\
        &ChatGen-Base(4B)           &0.236           &0.171          &0.448           &41.2          &69.4          &23.4          & -4.3  & 51.9    \\
        &ChatGen-Base(8B)           &0.252           &0.201          &0.462           &41.4          &70.6          &23.4          & -3.1   & 52.5  \\
        &\cellcolor{gray!22}ChatGen-Evo (2B)           &\cellcolor{gray!22}0.283           &\cellcolor{gray!22}0.231          &\cellcolor{gray!22}0.493           &\cellcolor{gray!22}40.7               &\cellcolor{gray!22}72.5              &\cellcolor{gray!22}25.0             & \cellcolor{gray!22}5.1      & \cellcolor{gray!22}59.2    \\
    \bottomrule
    \end{tabular}
  \label{tab:main_results}
\end{table*}

\begin{table}[!t]
\centering
\caption{Efficiency comparison. Training efficiency is measured in GPU hours, while inference is expressed as the seconds required to process each data. The training was conducted on 8 A100 GPUs, and inference was performed on a single A100 80GB GPU.}
\vspace{-3mm}
\label{tab:efficiency}
\begin{tabular}{cccc}
\toprule
Method   & Params  & Training & Inference\\
\midrule
ChatGen-Base(2B) & 2.21B & 76h &  1.1s\\
ChatGen-Base(8B) & 8.08B & 240h&  2.3s\\
ChatGen-Evo(2B) & 2.24B &  100h&  1.9s\\
\bottomrule
\end{tabular}
\end{table}

\vspace{-0.3cm}
\paragraph{Baseline Without Fine-Tuning.} We further establish a baseline that uses the default in-context learning capabilities of the MLLM for prompt rewriting, along with a single Stable Diffusion model and a fixed set of default parameters. This baseline helps emphasize the significance of prompt rewriting and multi-model utilization.


%

\subsection{Main Experiment}
\label{exp:results}

\subsubsection{Quantitative Results} 

Table \ref{tab:main_results} presents the main quantitative results of ChatGen-Evo compared to different baselines. Overall, ChatGen-Evo significantly outperforms other methods across all metrics, including both step-wise and final image quality evaluations. Specifically, the low performance of the baseline highlights the importance of effective prompt rewriting and multi-model selection, underscoring the necessity of dedicated Automatic T2I methods. Additionally, fine-tuning MLLMs with progressively larger parameter scales—from 2B to 8B—yields steady performance improvements. 
Remarkably, ChatGen-Evo achieves performance comparable to ChatGen-Base at 8B, despite utilizing a significantly smaller parameter scale of just 2B. 


The comparisons across different settings provide further insights. Prompt rewriting ability demonstrates strong generalizability, effectively transferring to rarely seen models. The higher few-shot performance can be attributed to the bias in data distribution between the two settings, as reflected in the "Baseline" results, which reveal a difference of approximately 0.3. In contrast, model selection and parameter configuration show a noticeable decline in performance under few-shot conditions, highlighting their reliance on more training samples. Therefore, exploring the scaling laws of prompt rewriting and enhancing model selection in few-shot scenarios are promising directions.



\begin{figure}[!t]
  \centering
   \includegraphics[width=1.0\linewidth]{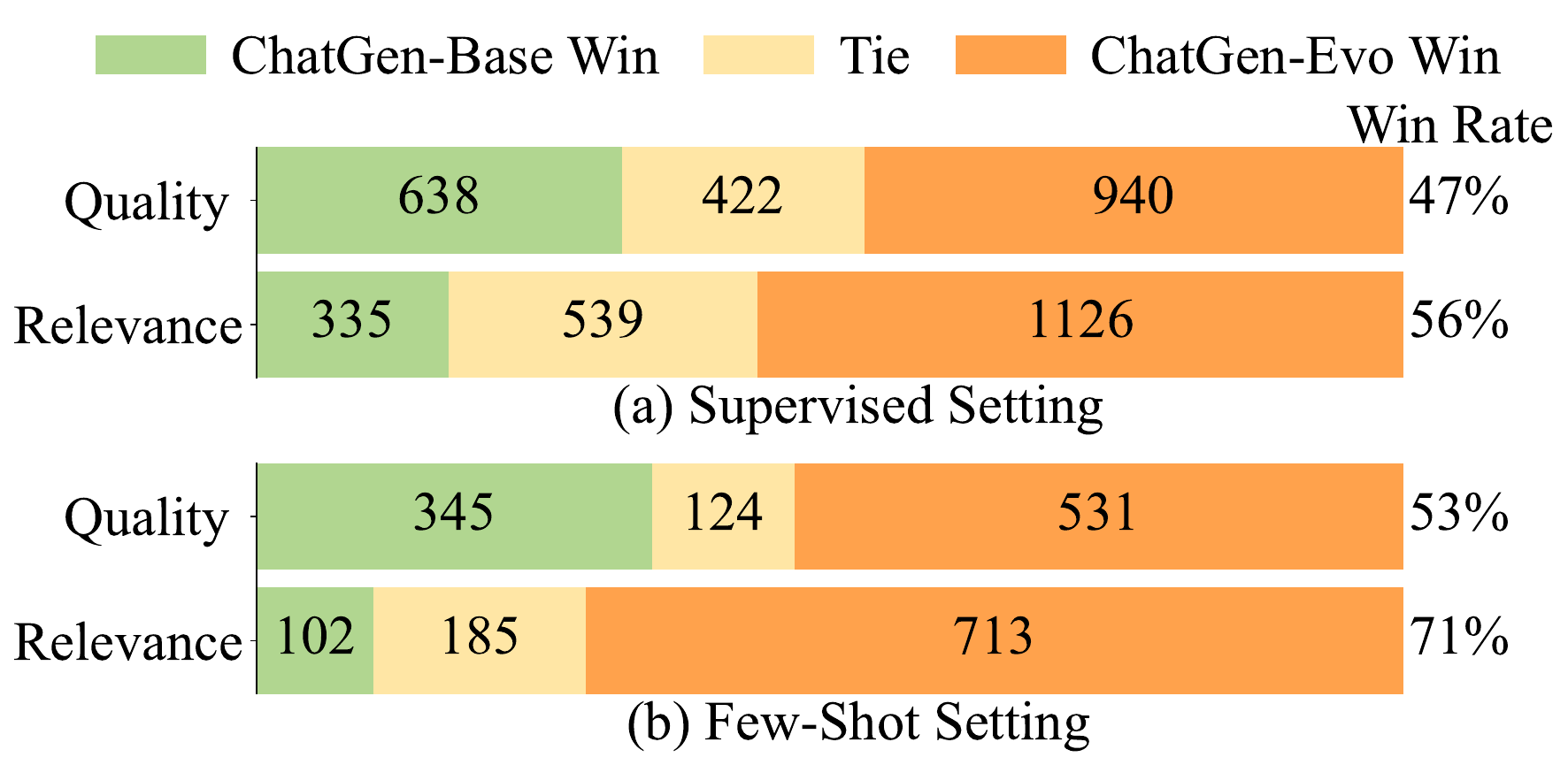}
   \caption{User study results of ChatGen-Base and ChatGen-Evo.}
   \label{fig:user_study}
\end{figure}
\subsubsection{Human Evaluation}
We conduct a user study using pairwise comparisons to further evaluate ChatGen-Base(8B) and ChatGen-Evo(2B). Users are presented with two images generated from the same input: one by ChatGen-Base and the other by ChatGen-Evo. It is tasked with selecting the image that better matches the image quality and relevance to the given input. We sample 2,000 image pairs for supervised and 1,000 for few-shot. 

As shown in Figure \ref{fig:user_study}, the human evaluation results are consistent with the quantitative metrics, highlighting that ChatGen-Evo outperforms ChatGen-Base in both image quality and relevance. Additionally, in few-shot settings, ChatGen-Evo demonstrates a higher win rate, showcasing the effectiveness of our approach in data-scarce scenarios.
\vspace{-0.4cm}
\subsubsection{Efficiency Comparison}

Similar to existing multi-stage reasoning methods~\cite{openai2024learning,wei2022chain,yao2024tree}, ChatGen-Evo does not offer an efficiency advantage over direct prediction approaches like ChatGen-Base. However, the significant performance gains more than compensate for this drawback. As shown in Table \ref{tab:efficiency}, ChatGen-Evo achieves the performance level of ChatGen-Base at 8B parameters using only 2B parameters. Therefore, when comparing efficiency under equal performance, ChatGen-Evo maintains a relative advantage over ChatGen-Base.

\begin{table}[!t]
  \centering
  \caption{Ablation experiments on the supervised setting.}
  \vspace{-3mm}
  \setlength{\tabcolsep}{3pt}
  \captionsetup{skip=5pt}
  \small
    \begin{tabular}{clc|cccc}
    \toprule
     Stage& Methods &\begin{tabular}[c]{@{}c@{}}Step\\Score $\uparrow$\end{tabular}& \begin{tabular} [c]{@{}c@{}}FID\\Score $\downarrow$\end{tabular}      & \begin{tabular}[c]{@{}c@{}}Image\\Reward $\uparrow$\end{tabular}  & \begin{tabular}[c]{@{}c@{}} Unified\\Metric $\uparrow$ \end{tabular} \\
    \midrule
        \multirow{3}{*}{\begin{tabular}[c]{@{}c@{}}Prompt \\ Writing\end{tabular}}
        & Baseline &0.026  & 19.3 &-13.1 & 50.8\\
        & Chat-Base & 0.184 & 21.3 & 3.1 & 59.7\\
        & Chat-Evo & 0.247 & 17.6 & 10.5 & 66.8 \\
    \midrule
        \multirow{2}{*}{\begin{tabular}[c]{@{}c@{}}Model \\ Selection\end{tabular}}
        & Chat-Base & 0.206 & 20.5 & 9.1 & 66.0 \\
        & Chat-Evo & 0.553 & 18.2 & 16.8 & 69.7 \\
    \midrule
        \multirow{2}{*}{\begin{tabular}[c]{@{}c@{}}Argument \\ Setting\end{tabular}}
        & Chat-Base & 0.384 & 17.9 & 10.4 & 69.1 \\
        & Chat-Evo & 0.871 & 17.1 & 17.9 & 70.3 \\
    \bottomrule
    \end{tabular}
  \label{tab:analysis}
\end{table}

\subsection{Analysis}
\label{exp:analysis}

\subsubsection{Capability Analysis.} 

We conduct ablation experiments on ChatGenBench to evaluate the contribution of individual steps to final performance by providing ground truth for the other steps.

\noindent \textbf{Prompt Writing:} We first compare the prompt writing capabilities of different methods, along with their final performance. In these experiments, each method’s predicted prompt was passed along with the correct model and argument. As shown in Table \ref{tab:analysis}, all methods exhibit noticeable gaps compared to human-validated prompts, emphasizing the complexity of prompt rewriting. Comparisons with the results in Table \ref{tab:main_results} also highlight the further improvements achieved by selecting the correct model and arguments, particularly for the ``Baseline" methods. Moreover, variations in prompts significantly influence the final results, highlighting its critical role in Automatic T2I.

\noindent \textbf{Model Selection:} Table \ref{tab:analysis} also shows the impact of model selection where human-validated prompt and argument are provided. With well-crafted prompts, ChatGen-Evo demonstrates a substantial performance boost from 32.8\% to 55.3\%, indicating the strong influence of prompt quality on model selection accuracy. This supports our perspective that Automatic T2I fundamentally involves multi-step reasoning. In contrast, ChatGen-Base, which directly predicts all results, fails to adapt and thus produces unchanged outputs. Furthermore, these results also emphasize the substantial impact of model selection on overall performance.

\noindent \textbf{Argument Configuration:} When provided with high-quality prompts and appropriate model selection, ChatGen-Evo exhibits notable performance improvements. It improves configuration accuracy from 53.7\% to 87.1\% and Unified Score from 65.9 to 70.3. Overall, while argument configuration has a relatively smaller impact compared to previous stages, it remains a crucial component.

The above findings suggest that outcomes in earlier steps significantly influence predictions in subsequent ones. Therefore, exploring more reasoning methods for advancing automatic T2I represents a promising research direction.

\begin{table}[!t]
  \centering
  \caption{The evaluation results of different input types.}
  \vspace{-0.3cm}
  \setlength{\tabcolsep}{6pt}
  \captionsetup{skip=5pt}
  \small
    \begin{tabular}{c|ccc|c}
    \toprule
         \multirow{3}{*}{\begin{tabular}[c]{@{}c@{}}Input \\ Type\end{tabular}} &\multicolumn{3}{c|}{\textbf{Step-wise Evaluation}} & \textbf{Final}  \\
     & \begin{tabular}[c]{@{}c@{}}Prompt\\BertScore $\uparrow$\end{tabular}      & \begin{tabular}[c]{@{}c@{}}Selection\\Acc $\uparrow$\end{tabular}  & \begin{tabular}[c]{@{}c@{}}Config\\Acc $\uparrow$\end{tabular}& \begin{tabular} [c]{@{}c@{}}Unified \\Score $\uparrow$\end{tabular}     \\
    \midrule
        Single & 0.252 & 0.331  & 0.539 & 68.1 \\
        M-Modal& 0.277 & 0.381 & 0.554 & 69.3  \\
        History& 0.165 & 0.259 & 0.505 & 60.1 \\
    \bottomrule
    \end{tabular}
  \label{tab:type_ana}
\end{table}

\subsubsection{Input Type Analysis.}
Table \ref{tab:type_ana} presents ChatGen-Evo performance across different input types. Multimodal inputs lead to better performance, as images may offer clearer prompt and model identification cues compared to text alone.  Additionally, handling historical data remains a significant challenge, with the lowest performance across all metrics. These results point to valuable directions in enhancing history-based prompt generation.

\begin{figure*}[]
  \centering
   \includegraphics[width=0.99\linewidth]{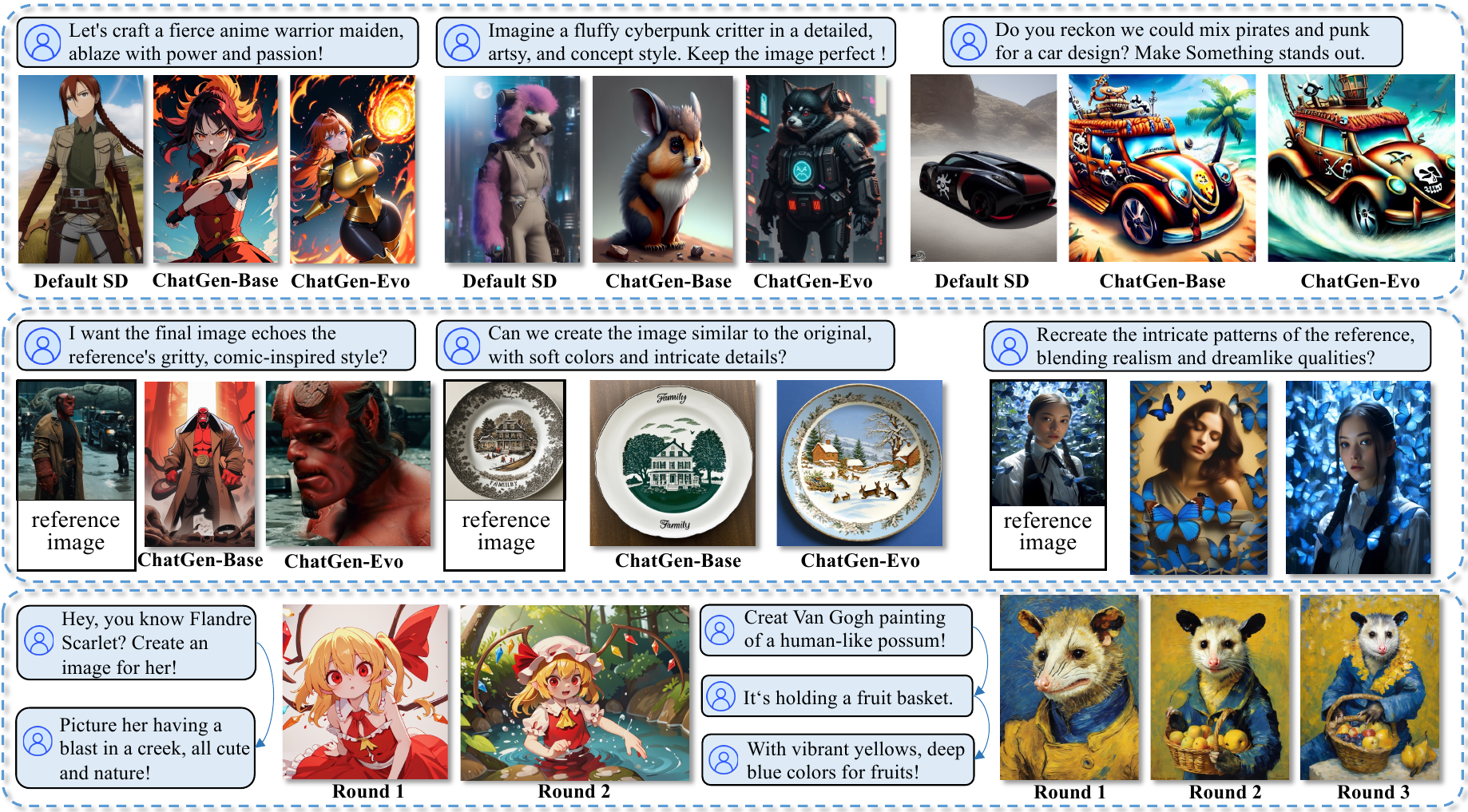}
   \caption{Examples of images generated by different methods. Three rows represent single, multi-modal and historical inputs, respectively.}
   \label{fig:visualization}
\end{figure*}
\subsection{Visualizations}
\label{exp:visualization}

\subsubsection{Qualitative comparisons}
Figure \ref{fig:visualization} presents examples of images generated by different methods. From the first row, it is evident that ChatGen-Evo understands user requirements and identifies suitable models to generate style-matching images. The second row demonstrates results based on multi-modal user inputs, where ChatGen-Evo shows a superior understanding of the reference image and preserves more details to generate refined outputs. The third row illustrates ChatGen-Evo's capability in handling historical data, ensuring that each step inherits the previous style while making appropriate modifications based on user requirements. 

\subsubsection{Qualitative results with step-wise outputs}

\begin{figure*}[t]
  \centering
   \includegraphics[width=1.0\linewidth]{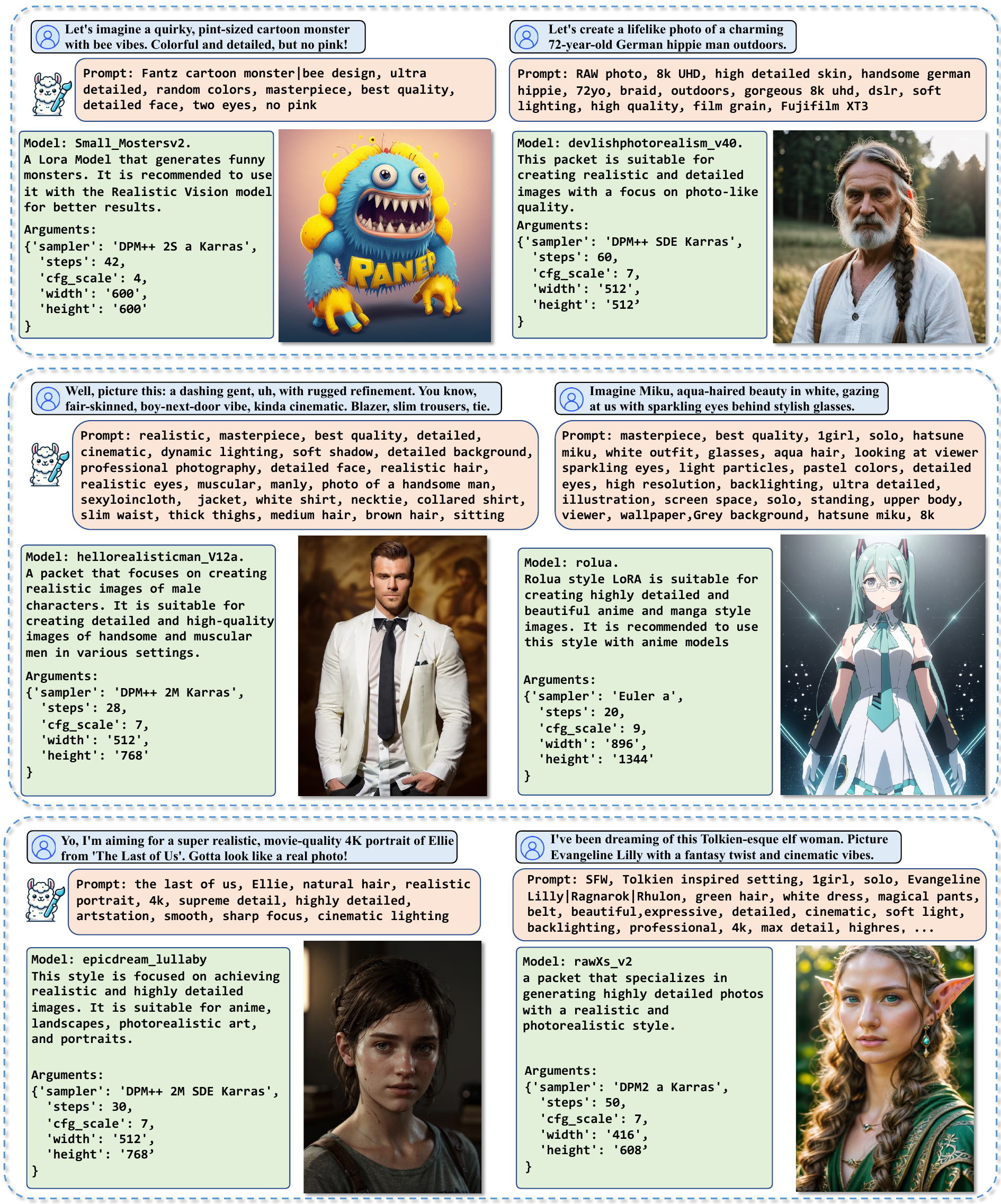}
   \caption{Examples of single inputs with step-wise outputs.}
   \label{fig:visualization_single}
\end{figure*}

\begin{figure*}[t]
  \centering
   \includegraphics[width=1.0\linewidth]{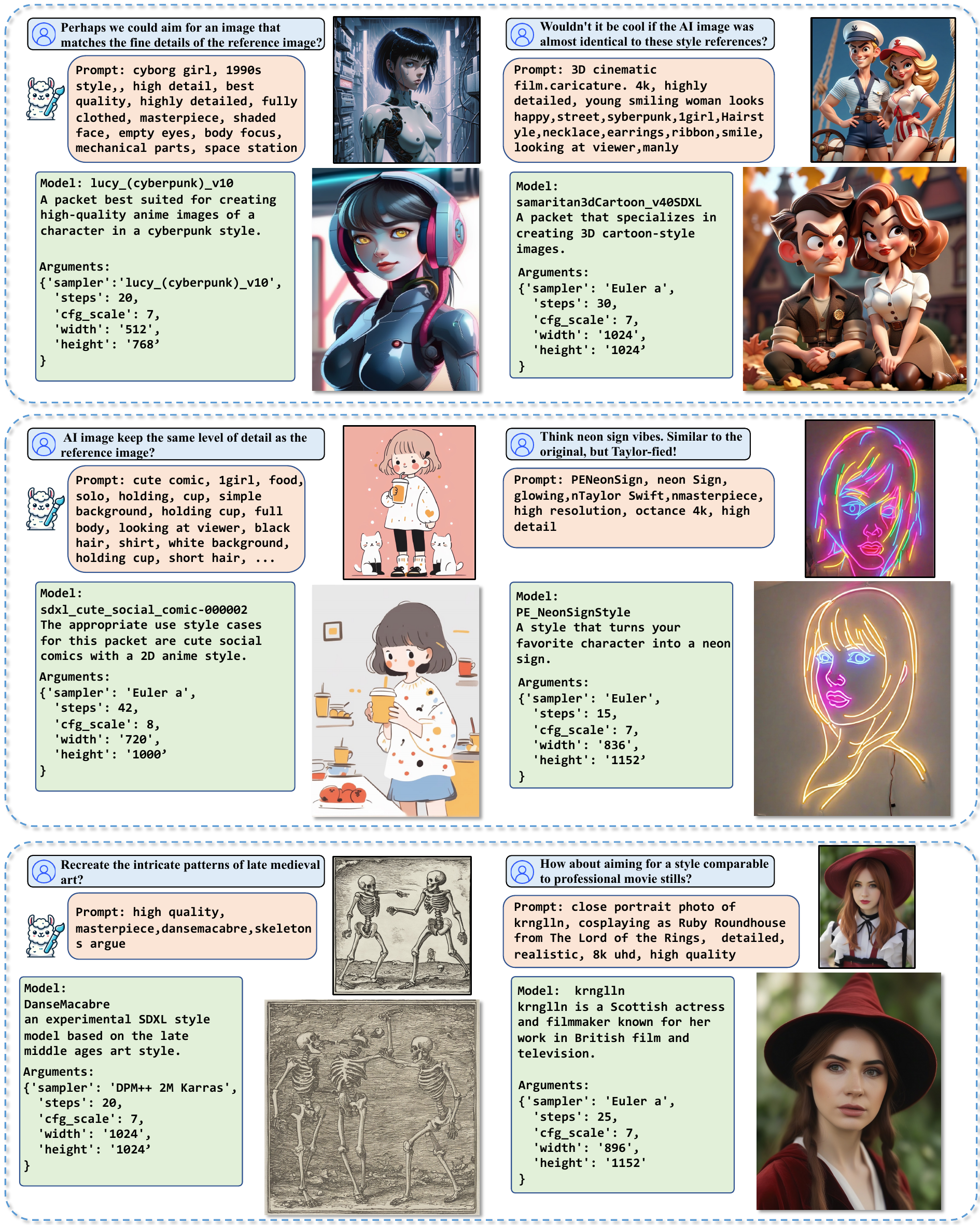}
   \caption{Examples of multimodal inputs with step-wise outputs. The image in the top-right corner represents the input reference image.}
   \label{fig:visualization_multi}
\end{figure*}

In Figures \ref{fig:visualization_single} and \ref{fig:visualization_multi}, we present the step-wise outputs and final images of ChatGen-Evo. It can be observed that ChatGen-Evo effectively rewrites high-quality professional prompts based on the user's freestyle input. Furthermore, ChatGen-Evo selects suitable models to match the user's desired style or character. Finally, it generates appropriate argument configurations to ensure the high quality of the resulting images. These high-quality images, produced through well-designed step-wise outputs, demonstrate the value of Automatic T2I. It relieves users from tedious steps and automates the production of desired images directly from their freestyle input.

\subsubsection{Comparisons with commercial models}

\begin{figure*}[t]
  \centering
   \includegraphics[width=0.98\linewidth]{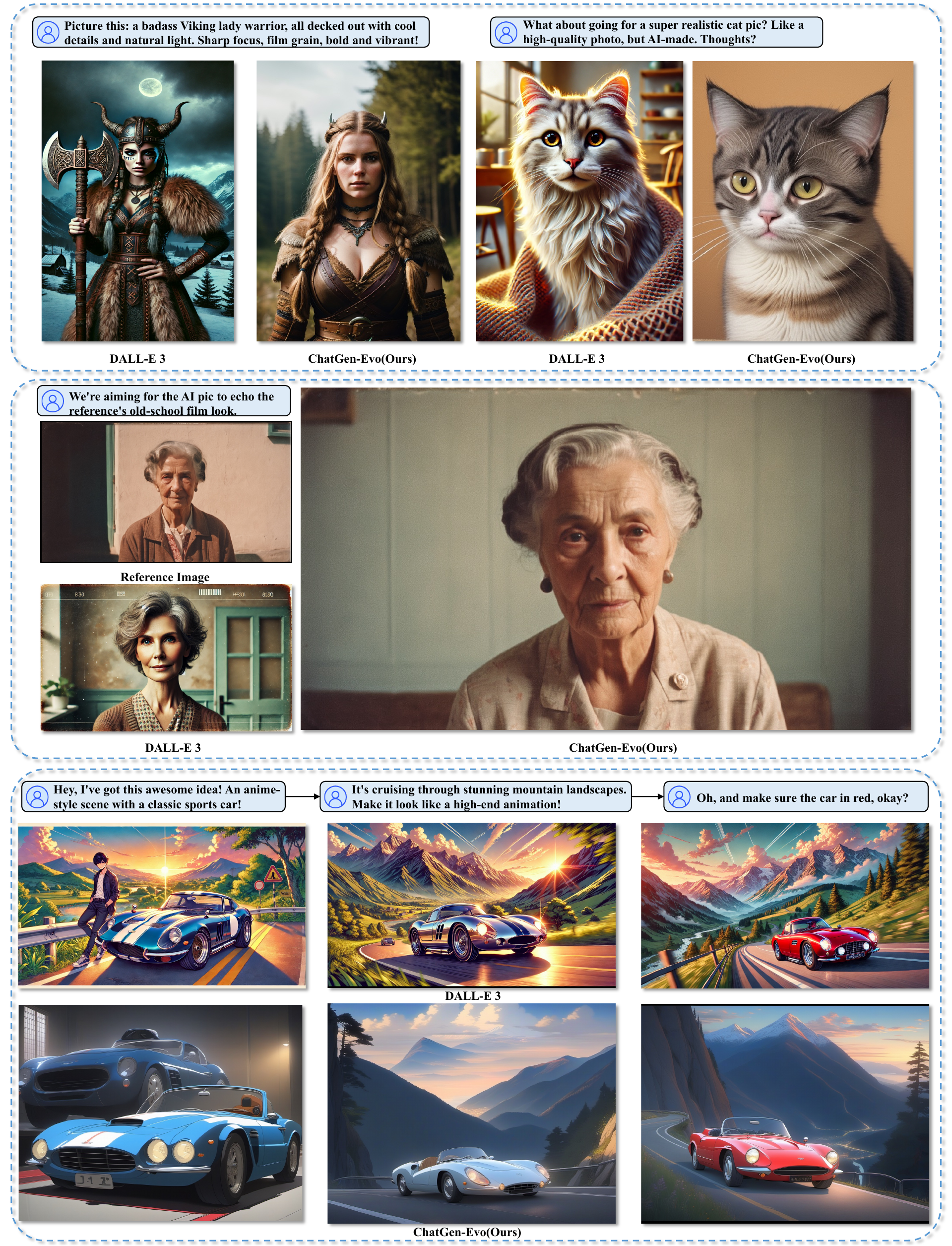}
   \caption{Examples of images generated by ChatGen-Evo and DALL-E 3. Three rows represent single, multi-modal and historical inputs, respectively.}
   \label{fig:visualization_dall}
\end{figure*}

In Figure \ref{fig:visualization_dall}, we compare the image quality of our method with the advanced commercial model DALL-E 3~\cite{betker2023dalle3}. While DALL-E is capable of generating high-quality images, its style is predominantly limited to a single type (anime-style). This limitation arises from its reliance on a single model, which cannot fully accommodate diverse and personalized styles. This highlights the value of our approach, which performs significantly better in scenarios requiring realistic styles or other personalized outputs. Additionally, we compare the results for history inputs. While DALL demonstrates strong performance, our open-source project contributes to advancing research in this field and helps bridge the gap between commercial models and open-source solutions in these aspects.

\section{Conclusion}
Our research aims to automate tedious steps in T2I generation, allowing users to simply describe their needs in a freestyle chatting way. We introduce \textbf{ChatGenBench} for benchmarking Automatic T2I task. It includes high-quality paired data with diverse freestyle user inputs, enabling comprehensive evaluation across all steps. Furthermore, we argue that Automatic T2I should be regarded as a multi-step reasoning task. Consequently, we propose \textbf{ChatGen-Evo}, a multi-stage evolution strategy that progressively equips models with essential automation skills. Extensive evaluations not only demonstrate the superiority of ChatGen-Evo but also provide valuable insights for advancing Automatic T2I. We believe this represents a significant step toward the future of automated generative models.

\clearpage
\clearpage

\clearpage

\newpage
{
    \small
    \bibliographystyle{ieeenat_fullname}
    \bibliography{main}
}



\end{document}